\documentclass[letterpaper, 10 pt, conference]{ieeeconf} 
\IEEEoverridecommandlockouts                             
\usepackage{color}

\newcommand\edits[1]{\textcolor{black}{#1}}
\usepackage{graphicx} % for pdf, bitmapped graphics files
\usepackage{mathtools} 
\usepackage{siunitx}

\title{\LARGE \bf Retraction of Soft Growing Robots without Buckling}

\author{Margaret M. Coad$^{1}$, Rachel P. Thomasson$^{1}$, Laura H. Blumenschein$^{1}$, \\Nathan S. Usevitch$^{1}$, Elliot W. Hawkes$^{2}$, and Allison M. Okamura$^{1}$
\thanks{$^{1}$Department of Mechanical Engineering, Stanford University, Stanford, CA 94305, USA {\tt\small mmcoad@stanford.edu}}%
\thanks{$^{2}$Department of Mechanical Engineering, University of California, Santa Barbara, Santa Barbara, CA 93106, USA}%
\thanks{This work was supported by National Science Foundation Award 1637446, Air Force Office of Scientific Research award FA2386-17-1-4658, and Toyota Research Institute (TRI). TRI provided funds to assist the authors with their research, but this article solely reflects the opinions and conclusions of its authors and not TRI or any other Toyota entity.}%
}

\begin{document}

\maketitle
\thispagestyle{empty}
\pagestyle{empty}

\begin{abstract}

Tip-extending soft robots that ``grow" via pneumatic eversion of their body material have demonstrated applications in exploration of cluttered environments. During growth, the motion and force of the robot tip can be controlled in three degrees of freedom using actuators that direct the tip in combination with extension. However, when reversal of the growth process is attempted by retracting the internal body material from the base, the robot body often responds by buckling rather than inverting the body material, making control of tip motion and force impossible. We present and validate a model to predict when buckling occurs instead of inversion, and we present an electromechanical device that can be added to a tip-extending soft robot to prevent buckling during retraction, \edits{restoring the ability of steering actuators to control the robot's motion and force} during inversion. Using our retraction device, we demonstrate three previously impossible tasks: exploring different branches of a forking path, reversing growth while applying minimal force on the environment, and bringing back environment samples to the base.

\end{abstract}

\section{Introduction}

\begin{figure}[tb]
      \centering
      \includegraphics[width=\columnwidth]{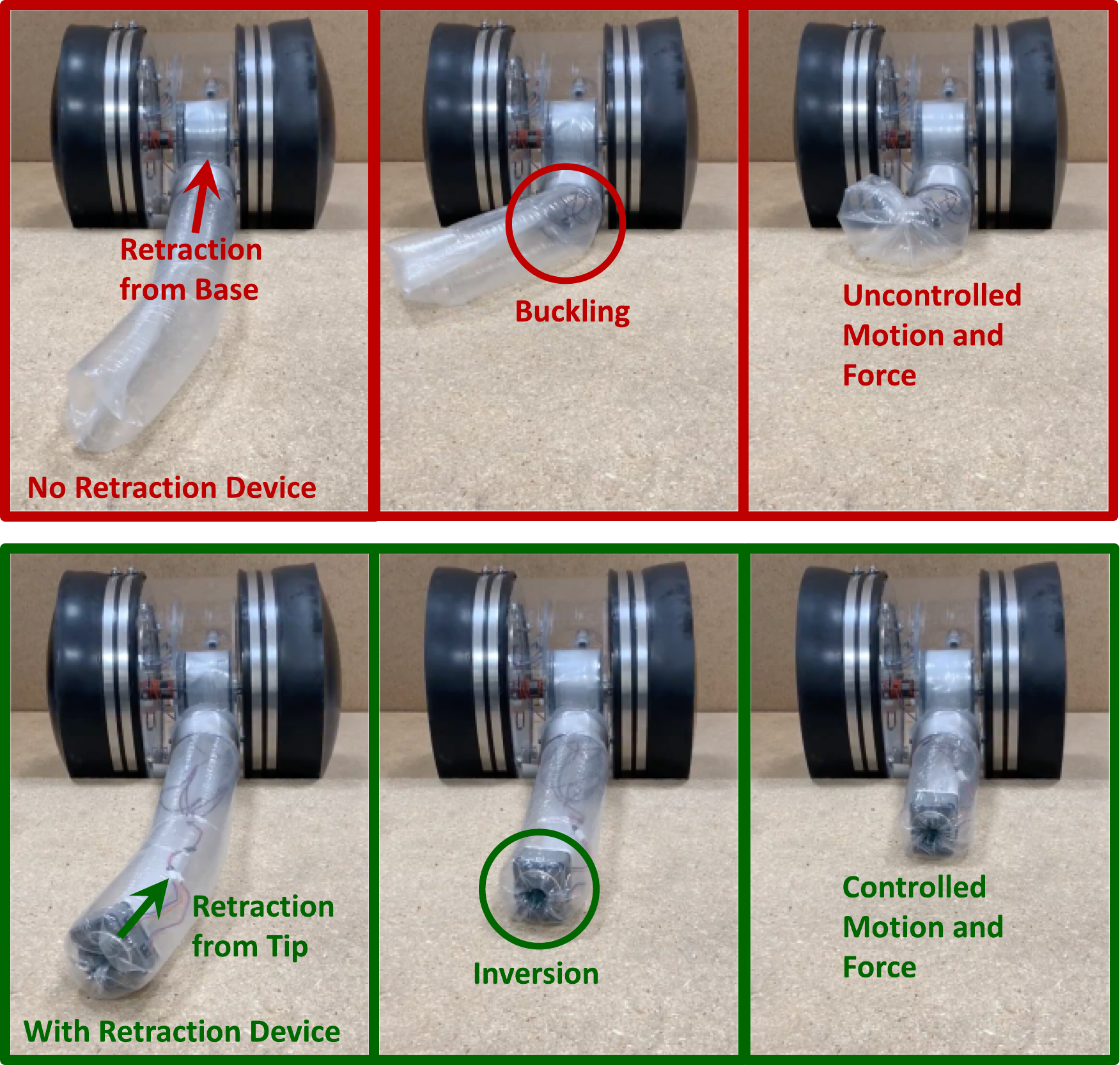}
      %\vspace{-0.1cm}
      \caption{Demonstration of a device to enable controlled retraction of a soft growing robot. After the robot grows from the base through eversion of its body material, retraction is attempted via tension applied on the robot tail with a motor in the base (top), and on the robot tail with a motorized retraction device at the tip (bottom). Without the device, the soft robot body often buckles, resulting in a lack of control over its motion and force. %With the device, the robot body retracts controllably through inversion of its body material.
      }
      \label{FirstFigure}
      \vspace{-0.5cm}
  \end{figure}

Pneumatically everting soft robots extend new material from their tip to navigate their environment, imitating plant-like growth~\cite{HawkesScienceRobotics2017, mishima2006development, tsukagoshi2011tip}. Movement through growth allows these robots to easily navigate cluttered environments by deforming around or through obstacles, moving independent of surface friction, and expanding their length many times over from a small form factor. Using actuation, like pneumatic artificial muscles, to asymmetrically shorten or lengthen the exterior surface of the soft robot body allows for \edits{motion and/or force application} in three degrees of freedom during growth~\cite{greer2018soft}. Pneumatically everting soft robots have been deployed for exploration and videography in an previously unexplored archaeological environment~\cite{CoadRAM2019}, and a hydraulic system was recently developed, showing promise for underwater applications such as exploration of coral reefs~\cite{luong2019eversion}.

\edits{Up to this point, the work on these robots has focused on their lengthening or \textit{growth}, instead of reversing growth, i.e. \textit{retraction}.} Control of the motion of the robot and the forces it exerts on the environment is difficult during retraction, because the soft robot body tends to buckle, especially after having grown to long lengths or into curved shapes. Adding the ability to \edits{controllably reverse growth would enable the steering actuators already used during growth to control the motion and forces applied by the robot during retraction, opening} up new capabilities for navigation and teleoperation. Controlled force during retraction enables removal of the robot without damage to delicate environments, and inversion of material at the tip can create a grasping behavior, allowing the robot to engulf objects encountered along its entire body.

In this paper, we explain the challenges associated with retraction, present a model predicting when controlled retraction is and is not possible, and present the design of a device  (Fig.~\ref{FirstFigure}) to aid in controlled retraction. We then demonstrate three new behaviors made possible by controlled retraction of pneumatically everting soft robots. The main contributions of this work are: (1) a model to predict when buckling and inversion occur during retraction, and (2) a device to expand the conditions under which retraction without buckling is possible. The work presented here is also useful for everting toroidal robots~\cite{orekhov2010mechanics, orekhov2010actuation}, and differs from prior work studying buckling due to environmental loads~\cite{godaba2019payload}.

\section{Problem Statement}

Two behaviors can occur when attempting to retract a pneumatically everting soft robot \edits{in free space} by pulling back on the inner body material from the base: ``inversion'' and ``buckling.'' During inversion, the outer robot body material (the ``wall'') inverts back into the deployed body at the tip and becomes the new inner robot body material (the ``tail''). This causes the robot tip to move in the direction opposite growth, as desired. In contrast, during buckling, the wall folds over on itself, allowing the tail to be pulled towards the base without inversion and causing the tip \edits{to move laterally, resulting in an unpredictable shape}. These two behaviors are shown in Fig.~\ref{FirstFigure}. \edits{The soft robot body can be kept from buckling due to the environment, but this will cause the robot to apply potentially undesirable lateral forces to the environment during retraction \cite{luong2019eversion, abrar2019epam}.}

\edits{With the addition of actuators such as cables or pneumatic artificial muscles \cite{greer2018soft,CoadRAM2019} that reversibly shorten different sides of the wall of the robot body, the robot tip position and the force exerted on the environment can be controlled in three degrees of freedom during growth. Steering actuators control the two degrees of freedom of lateral movement and/or force application of the robot tip, and eversion controls the third degree of freedom independent from steering. However, when the robot body buckles, the tail shortens faster than the body and places an additional length constraint on the robot shape, coupling the growth/retraction degree of freedom and the steering actuation. This causes undesired motion and/or force application in the lateral direction, rather than in the direction reversing growth.}% Thus, buckling during retraction causes lack of control of the tip motion and force.}

If sufficient force can be exerted from the robot base to retract the soft robot after buckling, it is possible to pull the soft robot body into the base in an uncontrolled manner %(first buckling and then inverting the buckled robot) 
and then start over with controlled growth from zero length. An analysis of the force required to retract such a curved robot is presented in~\cite{luong2019eversion}, and the force grows exponentially with the total angle formed by the path of the robot body~\cite{blumenschein2017modeling}.

The goal of this work is to understand why buckling during retraction occurs instead of inversion and to develop a means of ensuring inversion of the robot body under all conditions.

\section{Modeling and Experimental Characterization}

To understand how to prevent buckling during retraction, we measured retraction behavior using the base presented in~\cite{CoadRAM2019} and developed a model to predict whether a soft robot body will buckle or invert. Our model assumes that a robot of a given length, pressure, and curvature will either invert or buckle, depending on which behavior requires the lowest force applied on the tail. Because straight robots tend to buckle partway along their length during retraction (resembling axial beam buckling) and curved robots tend to buckle at the base during retraction (resembling transverse beam buckling), we model straight and curved robots independently. The following subsections describe the experiments and equations used to create the model.

\subsection{Tail Tension During Inversion}

\begin{figure}[t]
      \centering
      \includegraphics[width=.95\columnwidth]{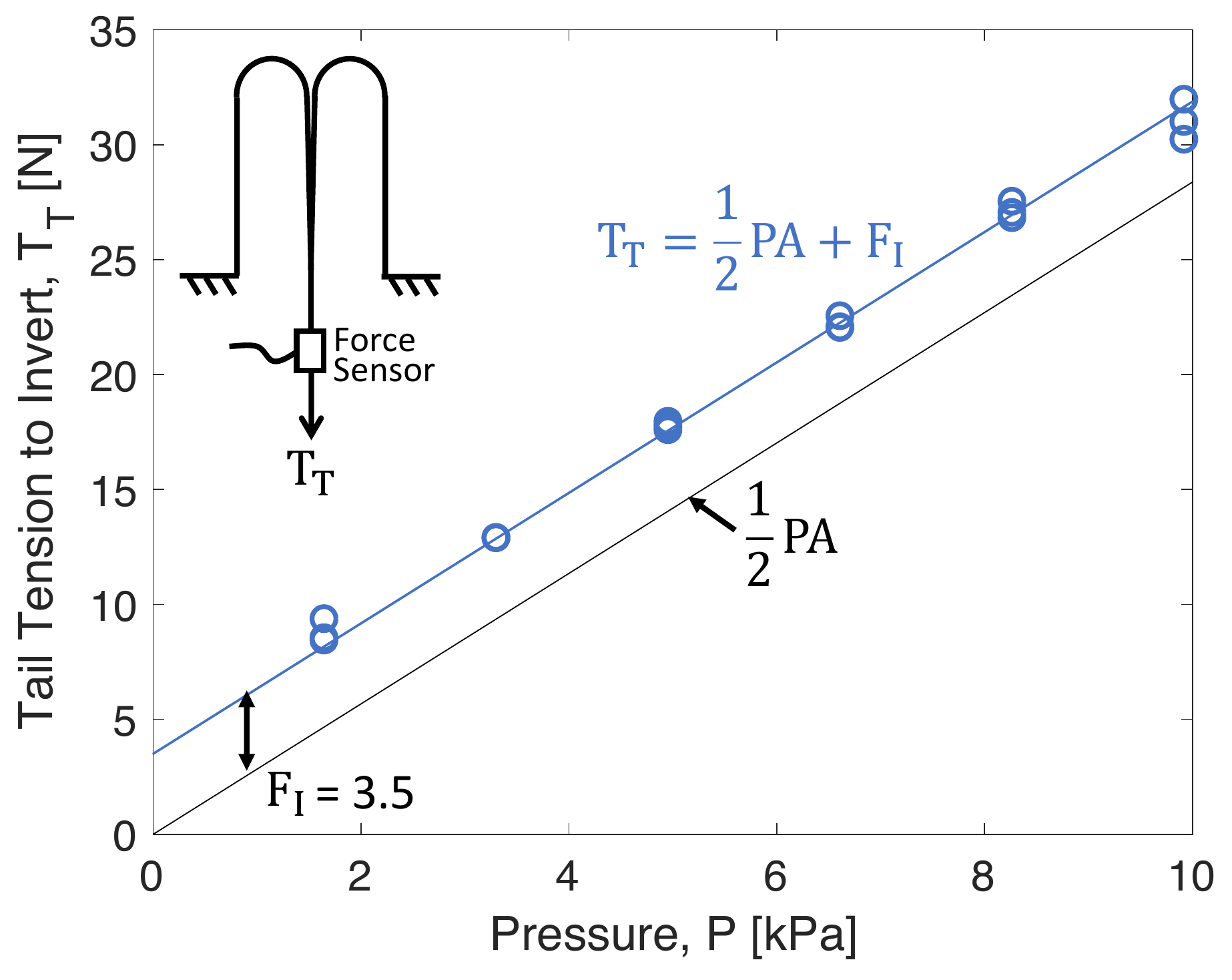}
      \caption{Measured tail tension $T_T$ required to invert the soft robot body at various pressures. The experimental setup is diagrammed in the top left. The blue line is the modeled tail tension (Eqn.~\ref{tail_tension}). As predicted by the model, tail tension required to invert a given soft robot body increases linearly with pressure, with a slope equal to half of the cross-sectional area of the robot body, $A/2$, and a pressure-independent offset inversion force $F_I$ that likely depends on the robot body material properties, diameter, and thickness.}
      \label{ForceToInvert}
      \vspace{-0.3cm}
  \end{figure}

During inversion, tension in the tail and the wall resist the internal pressure together (Section IV.A.1). Tension in the tail is higher than tension in the wall due to the force needed to deform the soft robot body material at the tip as the material transitions from the wall to the tail. This material deformation force was reported in \cite{HawkesScienceRobotics2017} and characterized in \cite{blumenschein2017modeling} for growth but not retraction. Tension in the tail during inversion $T_T$ is a function of pressure $P$: 
\vspace{-0.1cm}
\begin{eqnarray}\label{tail_tension}
T_T = \dfrac{1}{2}PA + F_I,
\end{eqnarray}
where $A$ is the cross-sectional area of the soft robot body (i.e. $\pi R^2$, where $R$ is the soft robot body radius) and $F_I$ is the additional force in the tail due to material deformation at the tip. The exact value of $F_I$ likely depends on the material properties, diameter, and thickness of the robot body.

To validate our tail tension model, we mounted an ATI Nano17 force sensor in line with the tail and used a motor and spool in the base to invert a straight soft robot body at various lengths and pressures. Figure~\ref{ForceToInvert} shows the experimental setup. Throughout this paper the robot body material used was low-density polyethylene (LDPE) with inflated diameter 8.5~cm and thickness 74~\si{\micro\meter}. A closed-loop pressure controller kept the pressure within the soft robot body constant during each trial. The base motor voltage was slowly increased until inversion began, and \edits{the maximum force measured during steady-state inversion was recorded} for each condition. No length dependence of the tail tension to invert the robot was found when data was taken at constant pressure \edits{(between 0 and 10 kPa)}, varying lengths between 35 and 190~cm. Data (Fig.~\ref{ForceToInvert}) was also taken while varying pressure between 0 and 10~kPa for a robot between 35 and 45~cm long, with three trials for each of six pressures. As predicted by the model, the tail tension to invert is linear with the pressure with a slope equal to half of the cross-sectional area of the robot body and an offset force. A best fit line with slope constrained to $A/2$ yielded a value of \edits{3.5}~N for~$F_I$.

\subsection{Straight Robot Buckling Model}

\begin{figure}[tb]
      \centering
      \includegraphics[width=.9\columnwidth]{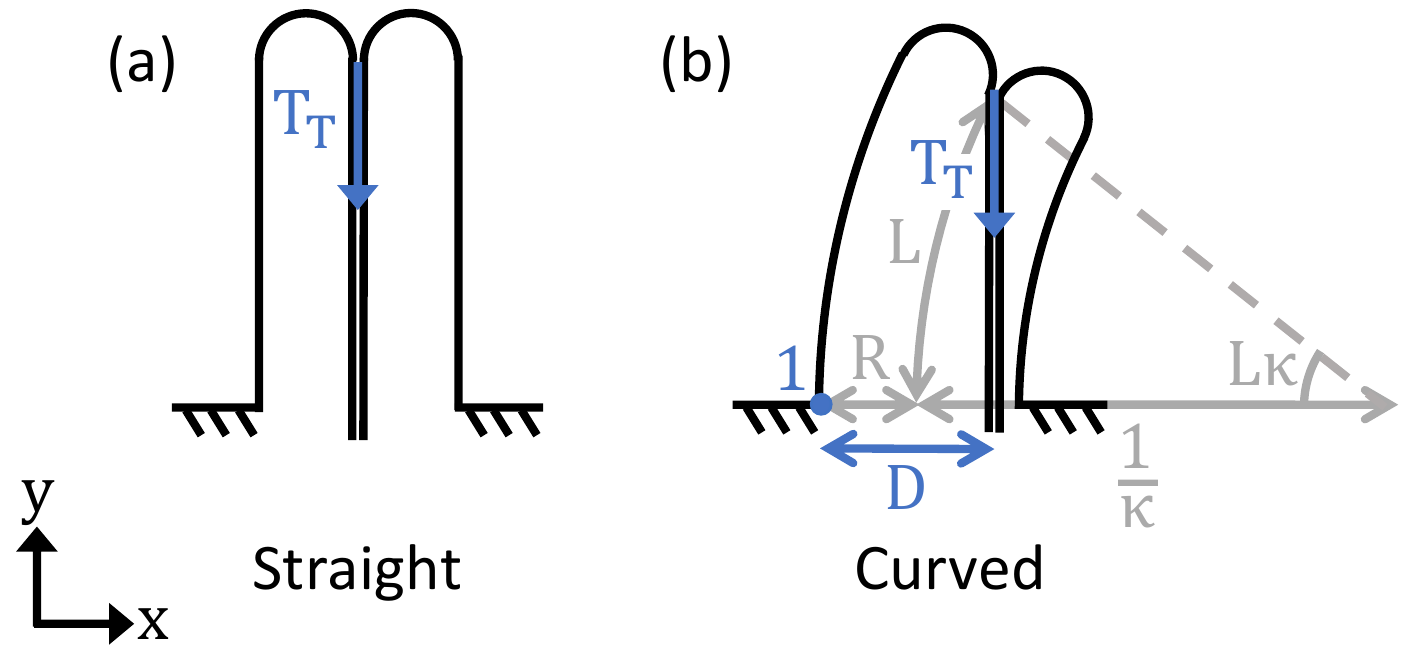}
      \caption{Modeling of buckling due to tail tension $T_T$. (a) For straight robots, the wall is an inflated beam undergoing an axial load due to the tail tension applied on the robot tip. (b) For curved robots, the wall is an inflated beam with a moment applied about Point 1 due to tail tension applied to the tip.}
      \label{StraightCurvedDiagram}
            \vspace{-0.3cm}
  \end{figure}

To determine when a straight robot body will buckle due to tail tension applied during retraction, we model the wall as an inflated beam on which the tail applies an axial force $T_T$ at the robot tip in the negative $y$ direction (Fig.~\ref{StraightCurvedDiagram}(a)). The force on the tip that causes buckling is found from the model presented in \cite{fichter1966theory} for axial buckling of an inflated beam:
\vspace{-0.1cm}
\begin{eqnarray}\label{axial_buckling}
F_{buckling} = \dfrac{E \pi^3 R^4 t P + E G \pi^3 R^3 t^2}{E \pi^2 R^2 t + R L^2 P + G t L^2},
\end{eqnarray}
where $E$ is the Young's modulus of the wall material, $t$ and $G$ are the thickness the shear modulus of the wall material, and $L$ is the length of the robot body. Values of 300~MPa for $E$ and 210~MPa for $G$ match those used for LDPE in~\cite{hammond2017pneumatic, Haggerty2019IROS}. This model is only valid when the wall is in tension. If the axial force is too large, the inflated beam collapses due to crushing~\cite{levan2005bending}:
\vspace{-0.2cm}
\begin{eqnarray}\label{axial_crushing}
F_{crushing} = PA.
\end{eqnarray}
For a given pressure and robot body length, if the tail tension required to invert the soft robot body is lower than both the axial buckling force and the axial crushing force, the robot will invert. Otherwise, the robot will buckle. Equating the tail tension to invert in Eqn.~\ref{tail_tension} with the crushing force in Eqn.~\ref{axial_crushing} and solving for pressure, we find that, for any length robot body, inversion is impossible at a pressure lower than 
\vspace{-0.1cm}
\begin{eqnarray}\label{minimum_pressure}
P_{min} = \dfrac{2F_I}{A},
\end{eqnarray}
which for our robot material and dimensions is 1.1 kPa.
Because the tail tension required for inversion is not dependent on length but the buckling force decreases with length, for each pressure above $P_{min}$, there is a critical length below which the soft robot body inverts and above which it buckles.

\subsection{Curved Robot Buckling Model}
Similar to the straight robot model, for a constant curvature robot, we consider the wall to be a constant curvature inflated beam on which the tail applies a force $T_T$ at the robot tip in the negative $y$ direction (Fig.~\ref{StraightCurvedDiagram}(b)), and we compare the tail tension to invert and to buckle the robot. In this case, the tail does not necessarily pass through the center of the robot body at the base (the tail is free to move laterally within the base in our hardware implementation), so it can create a net moment on the robot body, allowing transverse buckling to occur. We apply the same principle of transverse buckling presented in~\cite{leonard1960structural, comer1963deflections} to our curved inflated beam.

For any shape of robot body that has a cross-sectional area of $A$ at the base, the internal pressure applies a net moment on the wall about Point 1 of $PAR$ in the positive $z$ direction. The tail tension applies a moment on the wall about Point 1 of $T_TD$ in the negative $z$ direction, where $D$ is the moment arm of the tail tension force. Assuming the tail and wall connect at the center of the robot tip, the moment arm is
\vspace{-0.1cm}
\begin{eqnarray}\label{moment_arm}
D =  R + \dfrac{1}{\kappa}(1-\cos(L\kappa)),
\end{eqnarray}
where $\kappa$ is the robot body curvature and $L$ is the arc length of the centerline.
From a moment balance, the tail tension force required to buckle the curved robot body is
\vspace{-0.1cm}
\begin{eqnarray}\label{curved_buckling}
F_{buckling} = \dfrac{PAR}{D} = \dfrac{PAR}{R + \dfrac{1}{\kappa}(1-\cos(L\kappa))}.
\end{eqnarray}
This equation is only valid when the robot is not so curved or so long that the tail contacts the wall. However, the smallest moment arm ($D_{min}$) that causes buckling for a given curvature occurs before the tail contacts the wall. Equating the tail tension to invert in Eqn.~\ref{tail_tension} with the tail tension to buckle in Eqn.~\ref{curved_buckling}, $D_{min}$ is calculated as:
\vspace{-0.1cm}
\begin{eqnarray}\label{moment_arm_minimum}
D_{min} = \dfrac{PAR}{\dfrac{1}{2}PA+F_I}.
\end{eqnarray}

One caveat to our treatment of straight and curved robots separately: If our curved robot buckling model gives a longer transition length from inversion to buckling than the straight robot buckling model would give for the same pressure (which might happen for an extremely low curvature robot), the robot should be modeled as straight to match reality.

\subsection{Inversion and Buckling Data}

\begin{figure}[tb]
      \centering
      \includegraphics[width=\columnwidth]{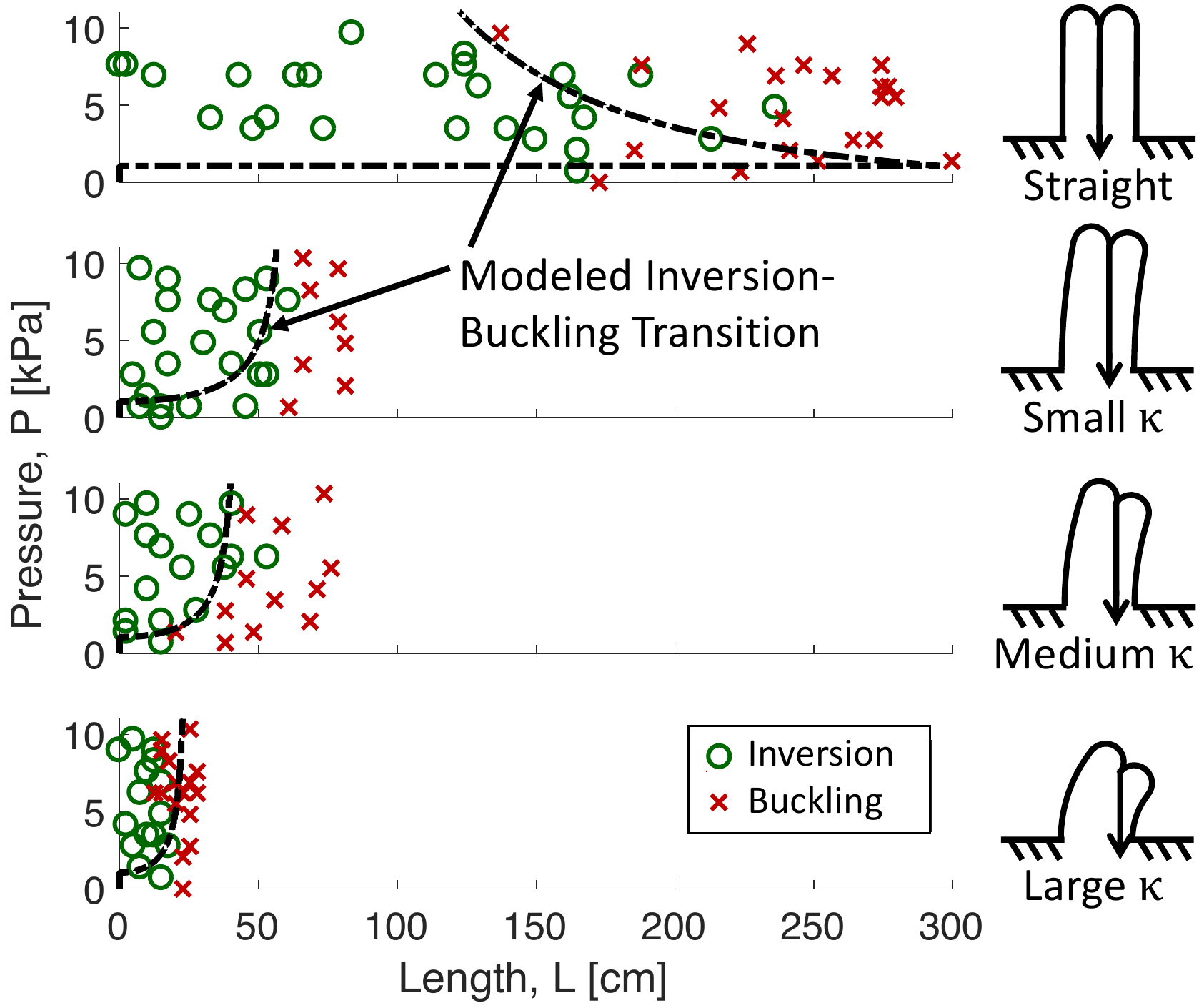}
      \caption{Experimental data showing behavior during retraction from the base across a range of pressures and lengths for different initial robot curvatures. \edits{Dotted} lines show modeled transitions from inversion to buckling. As predicted, the robot body inverts \edits{(green circles)} at shorter lengths and buckles \edits{(red x's)} at longer lengths. Also, more curved robots buckle at shorter lengths than straighter robots. Curved robots tested have radii of curvature of 455 cm (small $\kappa$), 225 cm (medium $\kappa$), and 72 cm (large $\kappa$).}
      \label{BuckleInvertScatter}
            \vspace{-0.3cm}
  \end{figure}

To validate the model, we experimentally retracted robots of four different curvatures at pressures between 0 and 10 kPa and lengths between 0 and 300 cm and observed whether they inverted or buckled. The curved robots were made by taping pinches in the wall at regular intervals, resulting in radii of curvature of 455 centimeters (small curvature), 225 cm (medium curvature) and 72 centimeters (large curvature). Data was taken after growing the robots horizontally on a flat surface. Fig.~\ref{BuckleInvertScatter} shows the results of this experiment, along with the modeled transition between inversion and buckling. As predicted, the robot body inverts at short lengths and buckles at long lengths and, the more curved the robot body, the shorter the transition length from inversion to buckling.

\section{Device Design}

Based on the modeling and experimental characterization of buckling and inversion presented in the previous section, we designed a device (Fig.~\ref{DeviceCAD}) to prevent buckling during retraction at any robot length, pressure, and curvature. For the purpose of our design, the key takeaway from the model is that at zero length, the soft robot body can always retract without buckling for any pressure above the minimum inversion pressure. The function of our device is to create an \textit{effective} length of zero for the purposes of retraction. The following subsections explain the forces involved in the function of our retraction device, its implementation, and an analysis of important design parameters.

\subsection{Working Principle}

\begin{figure}[tb]
      \centering
    \includegraphics[width=\columnwidth]{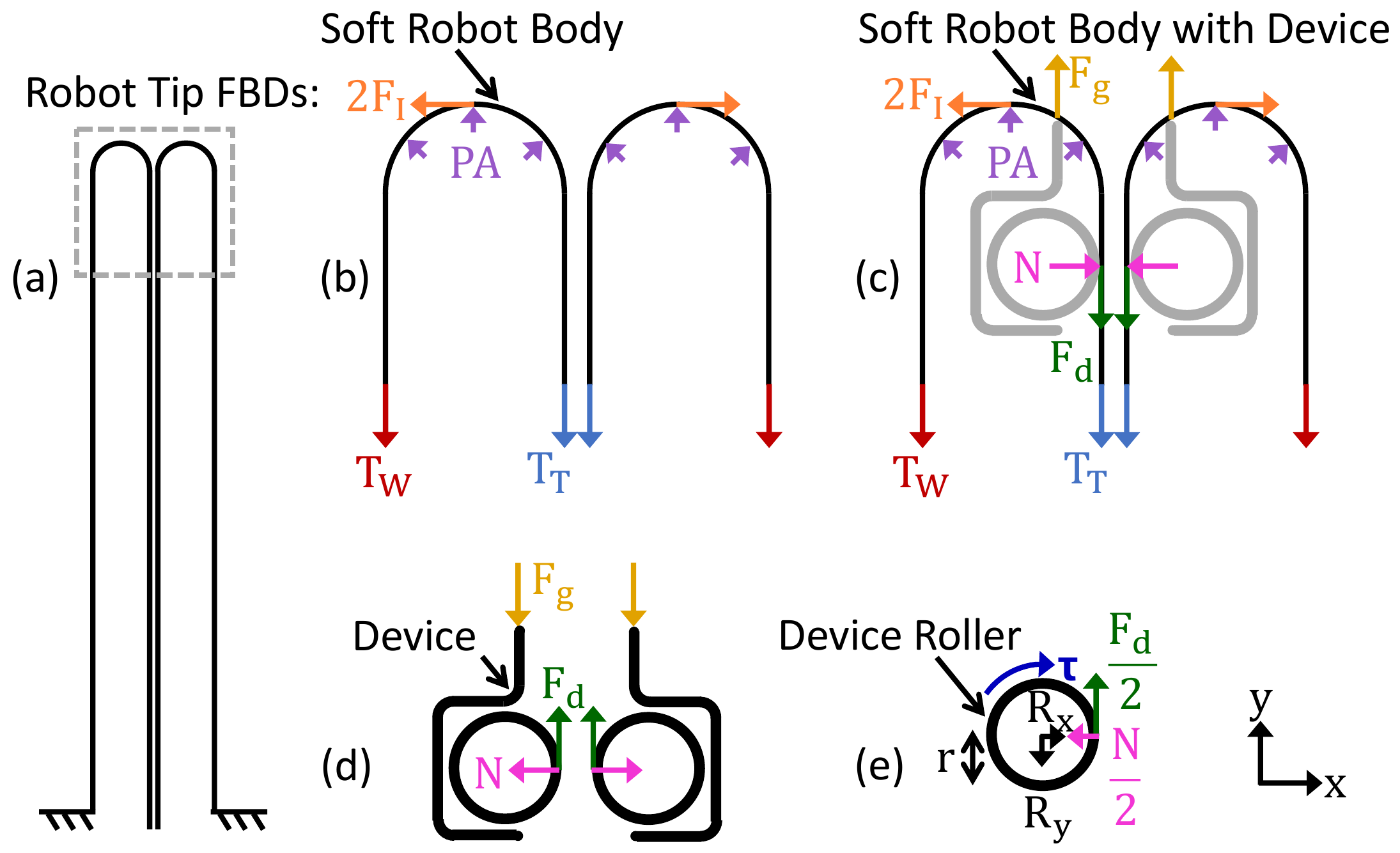}
      \caption{Free body diagrams (FBDs) showing the forces involved in inversion of the soft robot body with and without our retraction device. (a) All FBDs represent forces applied near the robot tip. FBDs are shown for (b) the soft robot body during inversion without the device, (c) the soft robot body when the device acts on it during inversion, (d) the device during inversion, and (e) one of the two rollers within the device during inversion. The device exerts a force $F_d$ on the tail of the soft robot body while providing an equal and opposite grounding force $F_g$ on the robot tip. Because the applied force is grounded to the robot tip, buckling due to this force is impossible, since the effective robot length for this force is zero. The device applies force through friction from a pair of motor-driven rollers that squeeze the tail.}
     \label{DeviceForces}
           \vspace{-0.3cm}
  \end{figure}

When retracting using the motor in the base, the motor and spool assembly applies the force on the tail and is grounded to the base, thus making the effective soft robot body length for the purpose of retraction the entire distance from base to tip. The distinguishing feature of our device is that it applies the force to retract on the tail \textit{while being grounded to the robot tip}, thus making the effective length of the robot zero.

\subsubsection{Forces without Device}
Figure~\ref{DeviceForces} shows free body diagrams relevant to the function of our retraction device. As shown in Fig.~\ref{DeviceForces}(a), all of the free body diagrams are drawn at the tip of the robot, and denote quasistatic forces during inversion. Fig.~\ref{DeviceForces}(b) shows the forces on the soft robot body during inversion without the device. Tension in the wall $T_W$ and tension in the tail $T_T$ both act in the negative $y$ direction. Pressure acts in all directions pointing outward from the inside of the soft robot body and producing a net force on the soft robot body of magnitude $PA$ in the positive $y$ direction. Additionally, an offset inversion force $F_I$ (multiplied by 2 to match Eqn.~\ref{tail_tension}) acts tangent to the wall material at the tip. The force balance in the $y$ direction gives: 
\begin{eqnarray}\label{force_body_no_device}
PA - T_W - T_T = 0 ,
\end{eqnarray}
and the tension balance along the soft robot body gives:
\begin{eqnarray}\label{force_cable_no_device}
T_W + 2F_I - T_T = 0 .
\end{eqnarray}
Solving Eqns.~\ref{force_body_no_device} and \ref{force_cable_no_device} for $T_T$, we get Eqn.~\ref{tail_tension}.

\subsubsection{Forces with Device}
Figure~\ref{DeviceForces}(c) shows the free body diagram for the soft robot body during inversion when the device is exerting forces at the tip. Our device creates three additional forces. The device's force on the tail $F_d$ adds to the tail tension in the negative $y$ direction while the device's grounding force $F_g$ is applied on the tip in the positive $y$ direction. Lastly, a normal force $N$ acts inward on the tail in the $x$ direction. The normal force arises from rollers that squeeze the tail to create the friction needed to apply the device force. Here $2F_I$ is the force to invert the soft robot body at zero pressure \textit{through the device} (characterized in Section IV.C). The force balance in the $y$ direction now yields
\begin{eqnarray}\label{force_body_device}
PA - T_W - T_T + F_g - F_d = 0 ,
\end{eqnarray}
and a tension balance along the soft robot body results in
\begin{eqnarray}\label{force_cable_device}
T_W + 2F_I - T_T - F_d = 0 .
\end{eqnarray}
Additionally, using the free body diagram of the device shown in Fig.~\ref{DeviceForces}(d), we sum the forces in the $y$ direction to see that the device and grounding forces are equal:
\begin{eqnarray}\label{force_device}
F_d - F_g = 0 . 
\end{eqnarray}
Solving Eqns.~\ref{force_body_device}, \ref{force_cable_device}, and \ref{force_device} for $T_T$, we have
\begin{eqnarray}\label{tail_tension_device}
T_T = \dfrac{1}{2}PA + F_I - \dfrac{1}{2}F_d ,
\end{eqnarray}
which indicates that the tail tension and the device force work together to balance the internal pressure and inversion force. If the device force increases, the tail tension necessary to invert the robot decreases. If the device applies enough force, we can invert the robot without applying any tail tension force, thus making buckling due to retraction impossible. Plugging $T_T = 0$ into Eqn.~\ref{tail_tension_device}, we solve for the device force required to invert the robot with zero tail tension:
\begin{eqnarray}\label{device_force_no_tail}
F_d = PA + 2 F_I . 
\end{eqnarray}

\subsubsection{Forces on Device Rollers}
Finally, Fig.~\ref{DeviceForces}(e) shows the free body diagram of one of the two rollers in our device. We implemented the device using a pair of motor-driven rollers that grip the tail using friction. The left roller (with radius $r$) experiences a torque $\tau$ in the negative $z$ direction, as well as half of the device force in the positive $y$ direction and half of the normal force in the negative $x$ direction. Additionally, it is supported by the device housing with reaction forces $R_x$ and $R_y$. Summing the moments on the roller in the positive $z$ direction about the center of the roller, we have
\begin{eqnarray}\label{roller_moment}
\dfrac{1}{2}F_d r - \tau = 0 ,
\end{eqnarray}
which results in
\begin{eqnarray}\label{roller_torque}
F_d = \dfrac{2\tau}{r} \leq \dfrac{2\tau_{max}}{r} ,
\end{eqnarray}
where $\tau_{max}$ denotes the maximum possible torque applied by the motor. Because the device force is applied using friction between the rollers and the tail (with coefficient of static friction $\mu_s$), we can also write
\begin{eqnarray}\label{roller_friction}
F_d \leq \mu_s N .	
\end{eqnarray}
Based on Eqns.~\ref{roller_torque} and \ref{roller_friction}, the amount of force that the device can apply is limited by the amount of torque that each motor can provide, as well as the friction coefficient and normal force between the rollers and the tail, \edits{and the normal force depends on the spacing between the rollers.}

\subsection{Implementation}

\begin{figure}[tb]
      \centering
      \includegraphics[width=\columnwidth]{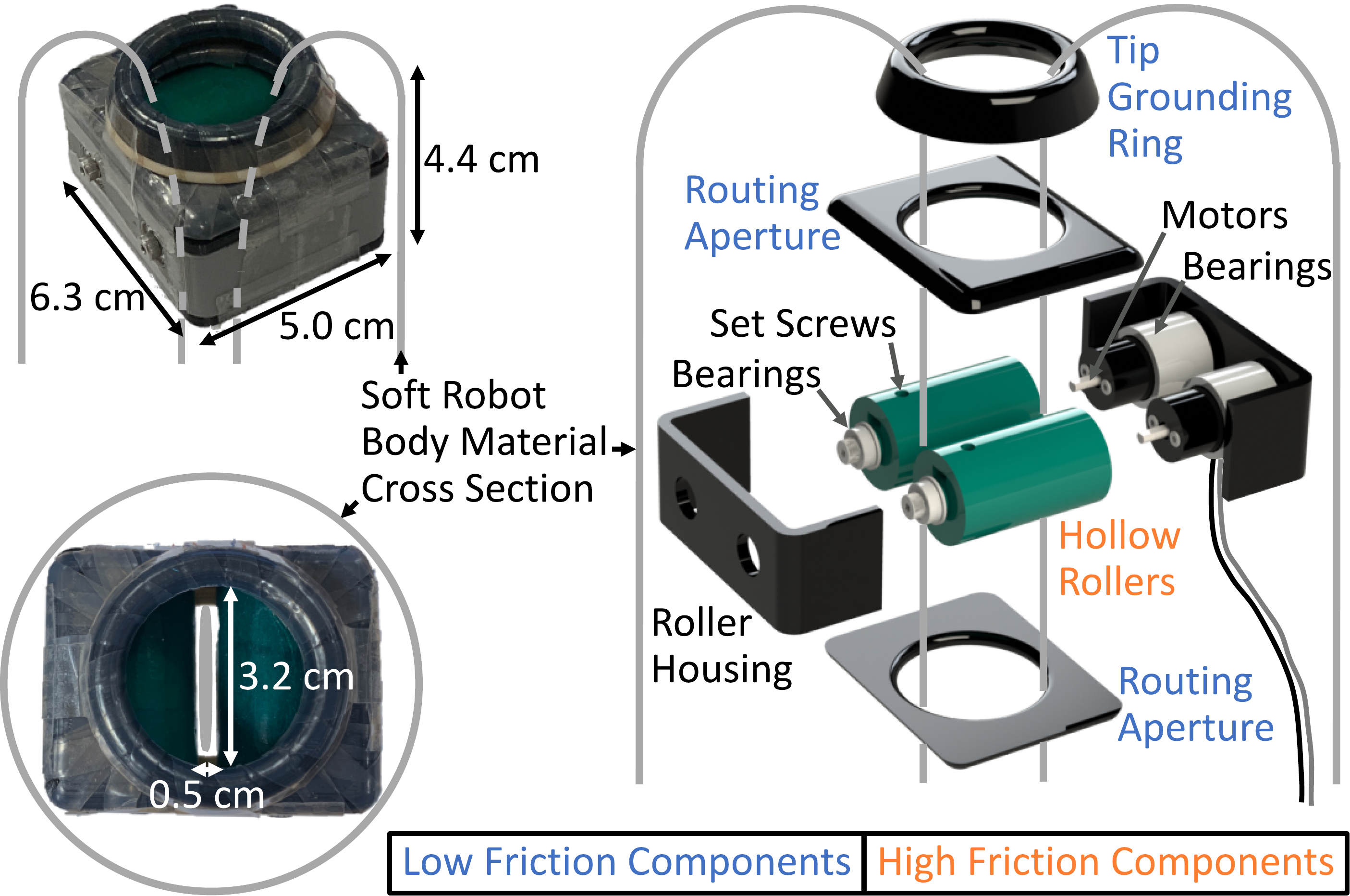}
      \caption{Implementation of our retraction device. The device is grounded to the tip of the soft robot body using a ring coated in low friction tape. It applies the force to invert the robot using a pair of rollers driven by motors and coated in high friction material. The soft robot body material is routed into the rollers using two routing apertures: one toward the robot tip and one toward the robot base. The motor wires run the length of the soft body.}
      \label{DeviceCAD}
            \vspace{-0.3cm}
  \end{figure}

Our retraction device is shown in Fig.~\ref{DeviceCAD}. Two motors (3070, Pololu Corporation, Las Vegas, NV) are mounted to one half of the 3D-printed roller housing so that the motor shafts protrude from holes in the housing. Two 3D-printed rollers fit over the motor housing and are rigidly connected to the motor shafts. The rollers roll on needle roller bearings mounted on the motor housing at one end and connect to the other half of the roller housing via ball bearings. The motor wires run internal to the soft robot body from the device to the base and through an airtight feedthrough in the wall of the base. The rollers are coated in a high-friction material (Non-Slip Reel, Dycem Corporation, Bristol, UK) to increase the device force that can be applied without slipping (Eqn.~\ref{roller_friction}). \edits{When the device motors are run in the retraction direction, the device first drives along the tail away from the base until it contacts the tip, at which point it begins retracting the robot body while remaining at the tip. When the device motors are run in the growth direction, the device either remains at the robot tip or drives along the tail towards the base, depending on whether the device moves slower or faster than the tail is being pulled toward the tip by pressure-driven growth.}

In our implementation, the friction force between the rollers and the soft robot body is high enough that the motor torque in Eqn.~\ref{roller_torque} is the limiting factor for the force the device is capable of exerting on the tail. Based on the maximum continuous torque of the motors (24.5 N-cm) and the roller radius (1.2 cm), the theoretical maximum device force is 41 N, which corresponds to retracting our soft robot body with zero tail tension at any pressure up to 6.2 kPa (Eqn.~\ref{device_force_no_tail}). Realistically, there are torque losses in the transmission of the motor torque through the rollers, and the value of $F_I$ increases due to the device (see the next subsection), so the maximum pressure at which this device can retract the robot on its own is closer to 2 kPa. \edits{Note, this is only the maximum retraction pressure. The pressure used to grow can be significantly higher.} The maximum motor speed is 33 RPM, which gives a retraction speed for the tip of 2.1 cm/s.

To prevent the tail from wedging itself between the edges of the rollers and the motor housing, two 3D-printed routing apertures with circular cutouts are attached to the top and bottom of the roller housing. These apertures are coated with ultra-low-friction tape (6305A16, McMaster-Carr Supply Company, Elmhurst, IL) to ease sliding of the tail through the apertures. Finally, to smooth the surface that contacts the robot tip, a 3D-printed circular tip grounding ring coated in low friction tape is attached to the top routing aperture. 

We attempted to minimize the size and weight of the device given the motors used. The device measures approximately 6.3 cm by 5.0 cm by 4.4 cm and weighs 106 g. The smallest diameter soft robot body that this version of the device can fit inside is approximately 8.1 cm in diameter. Note that, because the device only needs to contact the tail and the tip of the  soft robot body, not the wall, the same device can invert soft robot bodies of much larger diameter.

During retraction of long robots using our device, tail material builds up between the robot tip and the robot base, so the motor in the base must be used to take up the slack. The force required from the base motor to take up the slack in the tail can be calculated using the models presented in \cite{luong2019eversion, Haggerty2019IROS} and is two orders of magnitude smaller than the force required without the device.

\subsection{Device Aperture Size Analysis}

\begin{figure}[tb]
      \centering
      \includegraphics[width=\columnwidth]{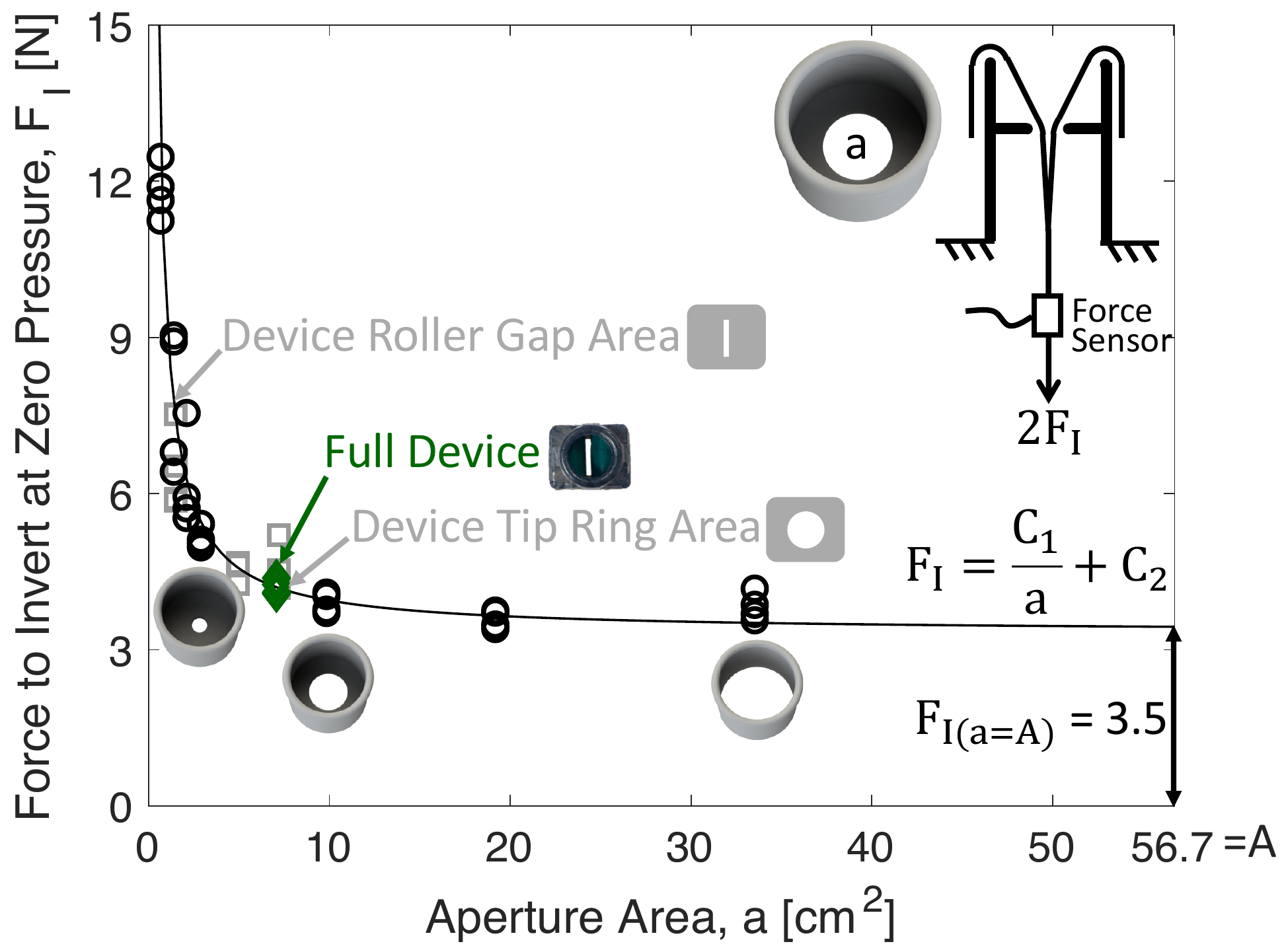}
      \caption{Measured force to invert the soft robot body, $F_I$, at zero pressure through various apertures coated in low friction tape and through our retraction device. Apertures were held fixed, and an uninflated soft robot body was manually pulled through each aperture with a force sensor in line with the tail. For circular apertures \edits{(black circles)}, force varies with the inverse of aperture area. \edits{Extrapolating the curve, an aperture area equal to the area of the soft robot body cross-section yields a force equal to the offset from Fig.~\ref{ForceToInvert}.} For \edits{rectangular apertures (grey rectangles),} force/area data points fall close to the same curve. The black line represents the resulting model of $F_I$, which is also presented in Eqn.~\ref{device_params}. Interestingly, the force to invert through our retraction device \edits{(green diamonds)} is approximately equal to the expected force to invert through a circular aperture with the area of the inside of its tip grounding ring and is much lower than the expected force to invert through the area of the gap between the rollers.
      }
      \label{DeviceParameters}
            \vspace{-0.3cm}
  \end{figure}
  
\begin{figure*}[tb]
      \centering
      \includegraphics[width=\textwidth]{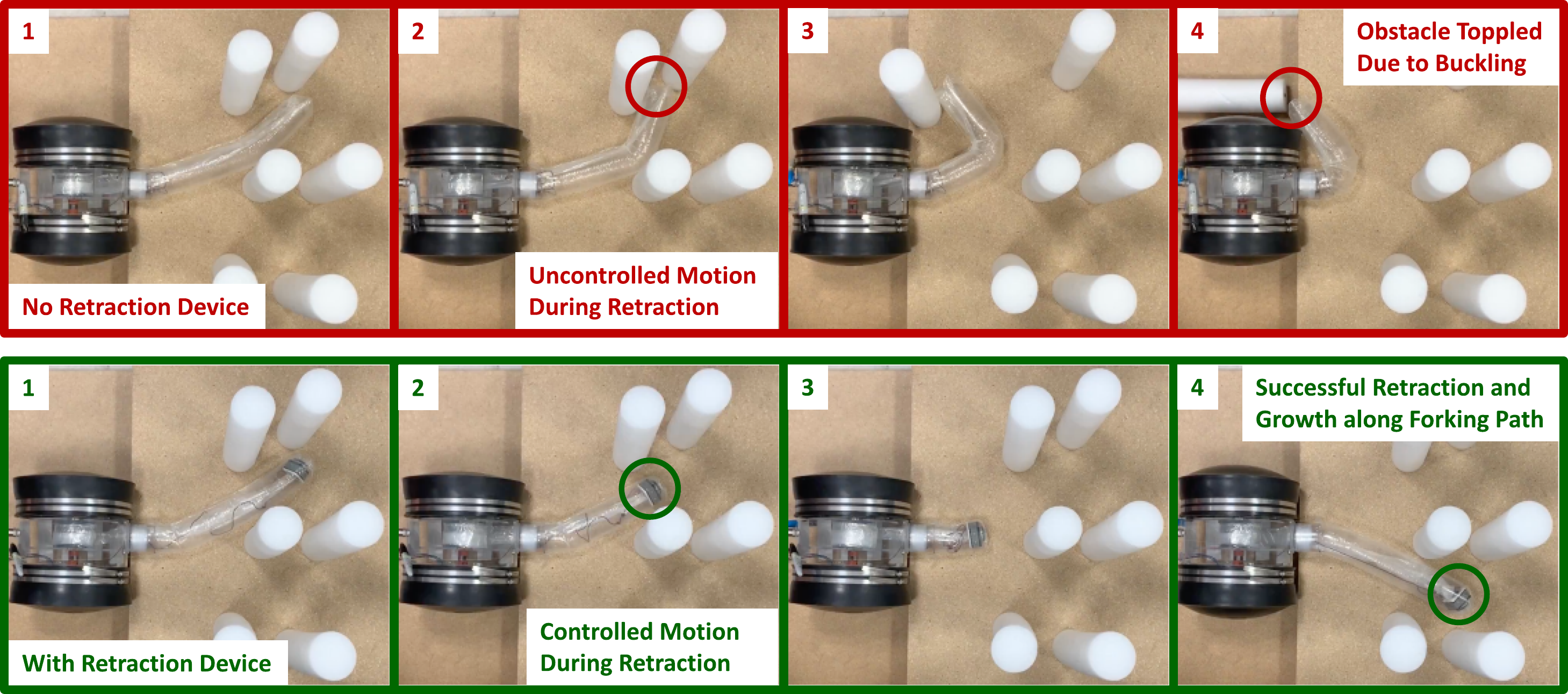}
      \caption{Controllably reversing growth after exploring one fork of a path to grow down another fork. The top time sequence shows that without the device, control of the tip position of the robot during retraction is impossible, leading to undesired environment contact and toppling one of the obstacles. The bottom sequence shows that with the device, the robot successfully navigates the forking path without moving any of the obstacles.}
      \label{NavigationDemo}
  \end{figure*}

The most important aspect of device geometry is the area of the smallest aperture through which the tail must slide during inversion through the device. \edits{This parameter determines the amount of force the device motors need to exert to begin inversion of the soft robot body at zero pressure.} To quantify the effect of this parameter, we measured the force required to invert the soft robot body through various apertures at zero pressure. Figure~\ref{DeviceParameters} includes a diagram of our experimental setup. The setup is almost the same as that described in Section III.A, except that the wall of the soft robot body is free to move and is not fixed to the base. Instead, a 3D-printed cylinder with a circular aperture inside is held fixed, and a soft robot body is manually pulled through the cylinder with the Nano17 force sensor in line with the tail. In this case, since the wall is free to move, the measured force is $2 F_I$. Seven different circular aperture sizes were investigated, with four trials for each aperture size. The maximum measured force for each trial was recorded as that trial's data point. Based on the data, we formed a descriptive model of the dependence of $F_I$ on aperture area $a$:
\vspace{-0.1cm}
\begin{eqnarray}\label{device_params}
F_I = \dfrac{C_1}{a} + C_2 ,	
\end{eqnarray}
where $C_1$ and $C_2$ were determined using a best fit curve as 6.1 N-cm\textsuperscript{2} and 3.3 N, respectively. Thus, the force to invert through a circular aperture varies inversely with aperture area. \edits{Rectangular apertures were also measured, and their forces fell close to the best fit curve for circular apertures.}
  
\begin{figure}[tb]
      \centering
      \includegraphics[width=\columnwidth]{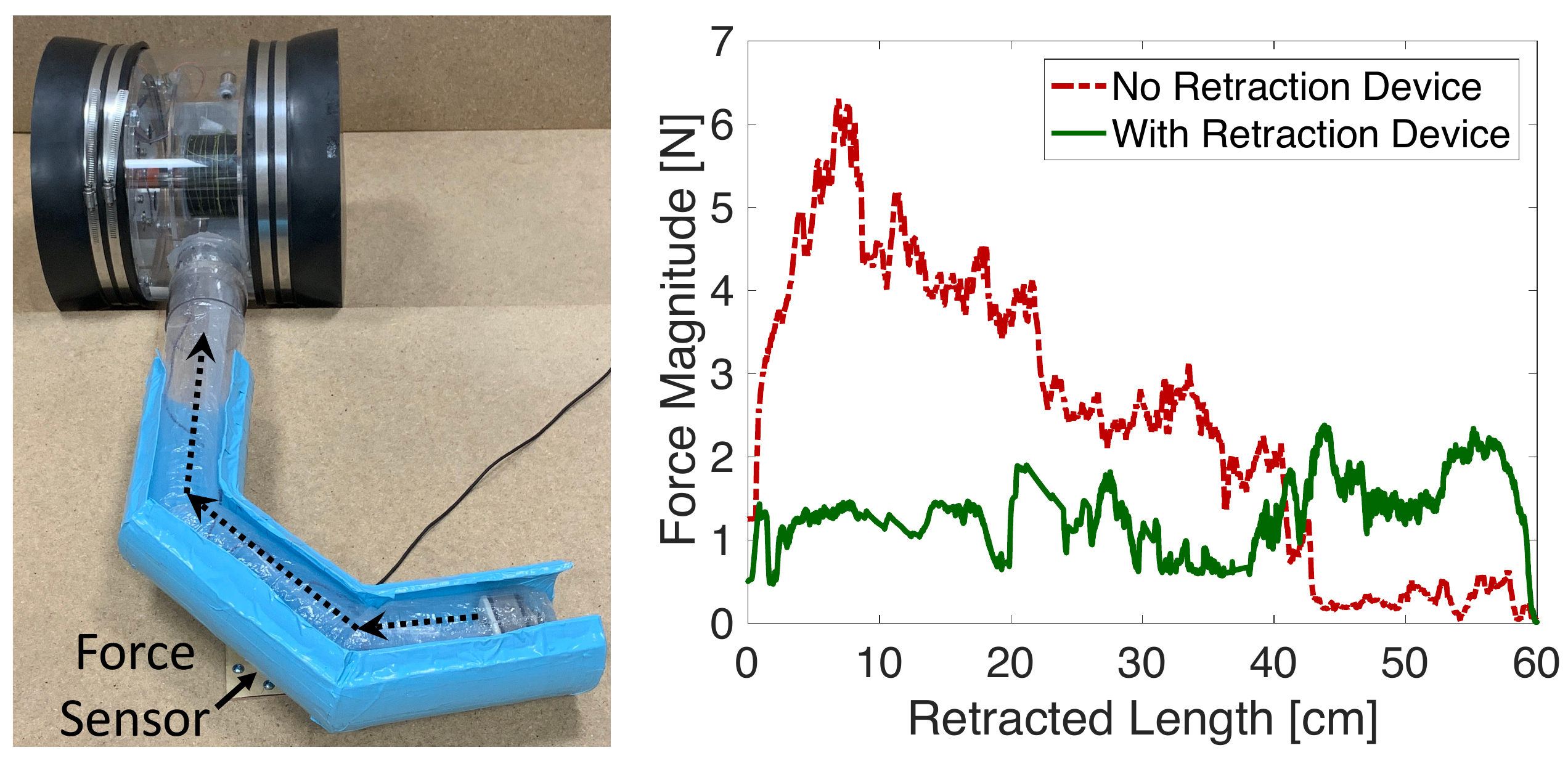}
      \caption{Retracting without exerting forces on the environment. Without the device, control of the force exerted on the environment is impossible, since the robot body braces itself against the environment to prevent buckling. With the device, the force exerted on the environment is much reduced and does not depend on the curvature or curved length of the robot body.}
      \label{ForceDemo}
            \vspace{-0.3cm}
  \end{figure}

We repeated the experiment with the \edits{unactuated} retraction device \edits{(with the roller bearing set screws disengaged from the motor shafts)} in place of the circular apertures. The force to invert through the device is approximately the same as the extrapolated force to invert through an aperture the size of the device's tip grounding ring (a 3.2 cm diameter circle) and significantly lower than the force to invert through a circular aperture the size of the rectangular gap between the rollers (0.5 cm $\times$ 3.2 cm). This indicates that, since the rollers roll along the tail, rather than making the tail slide through them, the force to invert through the device does not depend on the roller spacing but only on the smaller of the areas of the routing aperture and the inside of the tip grounding ring. According to the data, as long as the area of the smallest sliding aperture in the device is larger than approximately 5 cm\textsuperscript{2}, the force to invert through the device is minimal.

\section{Demonstration}

In this section, we demonstrate three new capabilities by comparing behaviors with and without the device.

\begin{figure*}[tb]
      \centering
      \includegraphics[width=\textwidth]{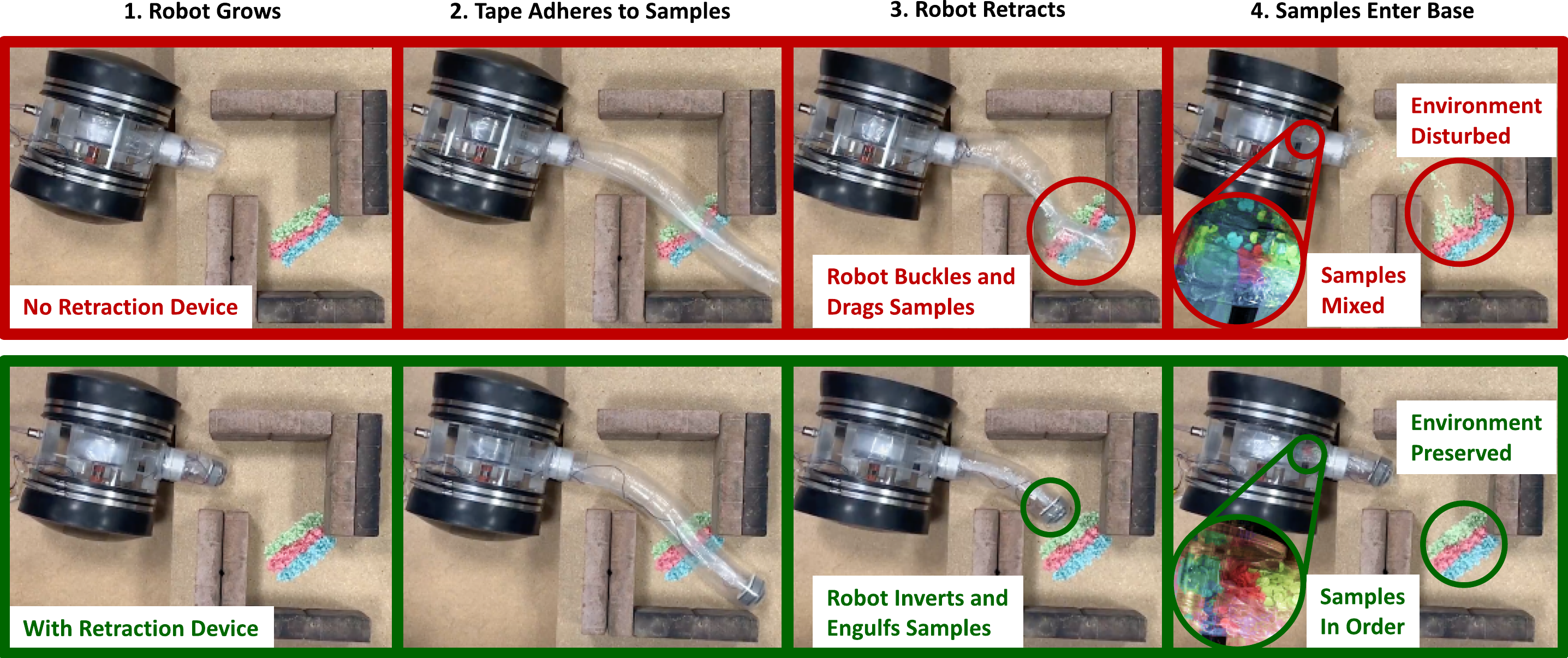}
      \caption{Growing over an environment and taking samples of the soil by engulfing them during retraction without disturbing the environment. The robot body has a layer of tape on the bottom to adhere to the samples. The top time sequence shows that without the device, the buckled robot body disturbs the environment and mixes up the samples (green, pink, and blue confetti) that are brought back to the base. The bottom sequence shows that with the device, the samples are brought back with an inherent record of how far along the robot they were found.}
      \label{SandEatingDemo}
  \end{figure*}

\subsection{Exploring a Forking Path}

One new capability for pneumatically everting soft robots is controlled position during retraction. A demonstration of this capability is shown in Fig.~\ref{NavigationDemo}, where sequential exploration of two different portions of a forking path is attempted. For this demonstration, the soft robot body was steered in one degree of freedom (left/right) by pulling on a pair of cables, one on the left and one on the right side of the soft robot body, running through a pair of tubular LDPE pockets taped along the entire length of the body. Both with and without the device, the soft robot body can be steered down one section of the forking path during growth. However, when retraction is attempted without the device (Fig.~\ref{NavigationDemo}, top) by pulling on the tail using the motor and spool in the base, the soft robot body buckles, and control of tip position during retraction is impossible. This results in one of the obstacles being toppled. In contrast, when retraction is attempted with the device applying the necessary force to invert at the robot tip (Fig.~\ref{NavigationDemo}, bottom), the robot tip's position can be completely controlled during retraction as well as growth, and the forking path can be successfully navigated without contacting any of the obstacles.

\subsection{Navigating a Delicate Environment}

A second new capability is controlled force during retraction. This capability is shown in Fig.~\ref{ForceDemo}, where the robot retracts after growing through a curved path \edits{while applying minimal force to the environment}. Here, the soft robot body grows inside a 9 cm inner diameter tube with two bends. \edits{The speed of growth and retraction is not directly controlled, but it is very slow (less than 1.7 cm/s).} \edits{The soft robot body is preformed to match the shape of the tube by pinching the wall at two points and taping over each pinch. This is to reduce the force between the soft robot body and the tube that holds it.} A Nano17 force sensor is mounted between the floor and the bottom of the tube. % and is the only connection point between the floor and the tube. 
During growth with and without the device, the force applied on the tube is minimal. During retraction with the device, the lateral force applied on the tube %in the plane of the curve 
is significantly lower than without the device (both at 1.4 kPa), and it does not depend on the robot length or curvature. Since the soft robot body would not buckle in free space during retraction with the device, it should not require support from the environment to avoid buckling. \edits{The measured force shown without the device (primarily directed into the curve) falls suddenly after 20 cm and 40 cm of retraction as the robot rounds each successive curve and requires less support from the environment to avoid buckling.} The measured force shown with the device is greater than zero likely due to imperfect alignment between the soft robot body and the tube. \edits{While the goal is zero force applied to the environment, other desired force values could be achieved during retraction with the device through the control of steering actuators attached to the soft robot body wall.}

\subsection{Environment Sampling}
A third new capability is environment sampling. This capability combines %two features of pneumatically everting soft robots: 
the ability to grow and retract over an environment without relative movement between the soft robot body wall and the environment with the ability to grasp and store objects by engulfing them during inversion. When tape is placed along the bottom of the soft robot body, it can stick to the environment as it grows (Fig.~\ref{SandEatingDemo}), and the soft robot body can pick up samples (green, pink, and blue confetti here) that then become packaged within the tail during inversion of the robot. The top of the figure shows that the environment is disturbed and the samples are mixed when this behavior is attempted without the device. The bottom of the figure shows that this sampling capability can be executed cleanly with our retraction device, where the soft robot body grows over the environment, retrieves the samples, and brings them back to the base. \edits{While this demonstration was conducted with a preformed robot, a similar capability could be achieved with a robot steered in such a way that the grown out portion does not move~\cite{HawkesScienceRobotics2017}.}

\section{Conclusion and Future Work}
We presented a model to predict when buckling and inversion occur during retraction for constant curvature pneumatically everting soft robots. A key takeaway from the model is that buckling due to retraction forces cannot occur if the robot body length between the force grounding point and the robot tip is zero. Using this insight, we presented the design and characterization of a new device to expand the conditions under which retraction without buckling is possible. We demonstrated several new robot behaviors made possible by our retraction device. The ability to control force and motion of the soft robot body during both retraction and growth improves the usability and practicality of pneumatically everting soft robots.

\edits{While the addition of our retraction device extends the capabilities of soft growing robots, it also limits some of their capabilities, at least in its current form. Firstly, the additional weight of the retraction device at the tip means that stronger steering actuators are needed to lift the robot tip against gravity, and the robot body is more likely to buckle due to its own weight when growing cantilevered. Secondly, the rigidity of the retraction device makes it impossible for the robot body to squeeze through gaps smaller than the size of the device. Thirdly, the growth and retraction speed of the soft robot body are limited by the speed at which the device motors can let out or pull in the tail material. Finally, the objects that can be engulfed by the soft robot body during retraction must be able to fit between the device rollers.}

Future work will incorporate the device into a complete pneumatically everting soft robot system through development of control algorithms to synchronize motion of the device motors with motion of the base motor and control of the pressure regulator. We also plan to explore grasping through inversion of larger objects and mounting sensors and other payloads at the tip of the robot.
%, potentially via a device with variable spacing between the rollers. Such a device would also facilitate retraction of robots with different types of actuators embedded on their body. Finally, we will explore ways to integrate the device with mounting of sensors and other payloads at the tip of the robot, and to use the device as a carriage to transport payloads up and down the body of the robot.

% \addtolength{\textheight}{-12cm}   

\bibliographystyle{IEEEtran}
\bibliography{library}

\begin{thebibliography}{10}
\providecommand{\url}[1]{#1}
\csname url@samestyle\endcsname
\providecommand{\newblock}{\relax}
\providecommand{\bibinfo}[2]{#2}
\providecommand{\BIBentrySTDinterwordspacing}{\spaceskip=0pt\relax}
\providecommand{\BIBentryALTinterwordstretchfactor}{4}
\providecommand{\BIBentryALTinterwordspacing}{\spaceskip=\fontdimen2\font plus
\BIBentryALTinterwordstretchfactor\fontdimen3\font minus
  \fontdimen4\font\relax}
\providecommand{\BIBforeignlanguage}[2]{{%
\expandafter\ifx\csname l@#1\endcsname\relax
\typeout{** WARNING: IEEEtran.bst: No hyphenation pattern has been}%
\typeout{** loaded for the language `#1'. Using the pattern for}%
\typeout{** the default language instead.}%
\else
\language=\csname l@#1\endcsname
\fi
#2}}
\providecommand{\BIBdecl}{\relax}
\BIBdecl

\bibitem{HawkesScienceRobotics2017}
E.~W. Hawkes, L.~H. Blumenschein, J.~D. Greer, and A.~M. Okamura, ``A soft
  robot that navigates its environment through growth,'' \emph{Science
  Robotics}, vol.~2, no.~8, p. eaan3028, 2017.

\bibitem{mishima2006development}
D.~Mishima, T.~Aoki, and S.~Hirose, ``Development of pneumatically controlled
  expandable arm for search in the environment with tight access,'' in
  \emph{Field and Service Robotics}.\hskip 1em plus 0.5em minus 0.4em\relax
  Springer, 2006, pp. 509--518.

\bibitem{tsukagoshi2011tip}
H.~Tsukagoshi, N.~Arai, I.~Kiryu, and A.~Kitagawa, ``Tip growing actuator with
  the hose-like structure aiming for inspection on narrow terrain.''
  \emph{IJAT}, vol.~5, no.~4, pp. 516--522, 2011.

\bibitem{greer2018soft}
J.~D. Greer, T.~K. Morimoto, A.~M. Okamura, and E.~W. Hawkes, ``A soft,
  steerable continuum robot that grows via tip extension,'' \emph{Soft
  Robotics}, pp. 95--108, 2018.

\bibitem{CoadRAM2019}
M.~M. Coad, L.~H. Blumenschein, S.~Cutler, J.~A.~R. Zepeda, N.~D. Naclerio,
  H.~El-Hussieny, U.~Mehmood, J.~Ryu, E.~W. Hawkes, and A.~M. Okamura, ``Vine
  robots: Design, teleoperation, and deployment for navigation and
  exploration,'' \emph{preprint arXiv:1903.00069}, 2019.

\bibitem{luong2019eversion}
J.~Luong, P.~Glick, A.~Ong, M.~S. deVries, S.~Sandin, E.~W. Hawkes, and M.~T.
  Tolley, ``Eversion and retraction of a soft robot towards the exploration of
  coral reefs,'' in \emph{IEEE International Conference on Soft Robotics
  (RoboSoft)}, 2019, pp. 801--807.

\bibitem{orekhov2010mechanics}
V.~Orekhov, M.~Yim, and D.~Hong, ``Mechanics of a fluid filled everting
  toroidal robot for propulsion and going through a hole,'' in \emph{ASME
  International Design Engineering Technical Conferences}, 2010, pp.
  1205--1212.

\bibitem{orekhov2010actuation}
V.~Orekhov, D.~W. Hong, and M.~Yim, ``Actuation mechanisms for biologically
  inspired everting toroidal robots,'' in \emph{IEEE/RSJ International
  Conference on Intelligent Robots and Systems}, 2010, pp. 2535--2536.

\bibitem{godaba2019payload}
H.~Godaba, F.~Putzu, T.~Abrar, J.~Konstantinova, and K.~Althoefer, ``Payload
  capabilities and operational limits of eversion robots,'' in \emph{Annual
  Conference Towards Autonomous Robotic Systems}, 2019, pp. 383--394.

\bibitem{abrar2019epam}
T.~Abrar, F.~Putzu, J.~Konstantinova, and K.~Althoefer, ``Epam: Eversive
  pneumatic artificial muscle,'' in \emph{2019 2nd IEEE International
  Conference on Soft Robotics (RoboSoft)}.\hskip 1em plus 0.5em minus
  0.4em\relax IEEE, 2019, pp. 19--24.

\bibitem{blumenschein2017modeling}
L.~H. Blumenschein, A.~M. Okamura, and E.~W. Hawkes, ``Modeling of bioinspired
  apical extension in a soft robot,'' in \emph{Conference on Biomimetic and
  Biohybrid Systems (Living Machines)}.\hskip 1em plus 0.5em minus 0.4em\relax
  Springer, 2017, pp. 522--531.

\bibitem{fichter1966theory}
W.~Fichter, ``A theory for inflated thin-wall cylindrical beams,''
  \emph{National Air and Space Administration (NASA) Technical Note}, 1966.

\bibitem{hammond2017pneumatic}
Z.~M. Hammond, N.~S. Usevitch, E.~W. Hawkes, and S.~Follmer, ``Pneumatic reel
  actuator: Design, modeling, and implementation,'' in \emph{IEEE International
  Conference on Robotics and Automation (ICRA)}, 2017, pp. 626--633.

\bibitem{Haggerty2019IROS}
D.~A. Haggerty, N.~D. Naclerio, and E.~W. Hawkes, ``Characterizing
  environmental interactions for soft growing robots,'' in \emph{2019 IEEE/RSJ
  International Conference on Intelligent Robots and Systems}, in press.

\bibitem{levan2005bending}
A.~Le-van and C.~Wielgosz, ``Bending and buckling of inflatable beams: Some new
  theoretical results,'' \emph{Thin-Walled Structures}, vol.~43, no.~8, pp.
  1166--1187, 2005.

\bibitem{leonard1960structural}
R.~W. Leonard, G.~W. Brooks, and H.~G. McComb~Jr, ``Structural considerations
  of inflatable reentry vehicles,'' \emph{National Air and Space Administration
  (NASA) Technical Note}, 1960.

\bibitem{comer1963deflections}
R.~Comer and S.~Levy, ``Deflections of an inflated circular-cylindrical
  cantilever beam,'' \emph{AIAA Journal}, vol.~1, no.~7, pp. 1652--1655, 1963.

\end{thebibliography}

\end{document}